\documentclass[10pt,twocolumn,letterpaper]{article}

\usepackage{ijcb}
\usepackage{times}
\usepackage{epsfig}
\usepackage{graphicx}
\usepackage{amsmath}
\usepackage{amssymb}

\usepackage[noend]{algpseudocode}
\usepackage{algorithmicx,algorithm}
\usepackage{makecell}
\usepackage{booktabs}
\usepackage{multirow}
\usepackage[misc]{ifsym}

\usepackage[pagebackref=true,breaklinks=true,letterpaper=true,colorlinks,bookmarks=false]{hyperref}

\ijcbfinalcopy 

\def\ijcbPaperID{147} 

\ifijcbfinal\fi
\begin{document}

\title{Domain Private and Agnostic Feature for Modality Adaptive Face Recognition}

\author{Yingguo Xu, Lei Zhang$^{(}$\textsuperscript{\Letter}$^)$, Qingyan Duan\\
School of Microelectronics and Communication Engineering, Chongqing University\\
Chongqing, China\\
{\tt\small \{xuyingguo, leizhang, qyduan\}@cqu.edu.cn}}

\maketitle
\thispagestyle{empty}

\begin{abstract}
Heterogeneous face recognition is a challenging task due to the large modality discrepancy and insufficient cross-modal samples. Most existing works focus on discriminative feature transformation, metric learning and cross-modal face synthesis. However, the fact that cross-modal faces are always coupled by domain (modality) and identity information has received little attention. Therefore, how to learn and utilize the domain-private feature and domain-agnostic feature for modality adaptive face recognition is the focus of this work. Specifically, this paper proposes a Feature Aggregation Network (FAN), which includes disentangled representation module (DRM), feature fusion module (FFM) and adaptive penalty metric (APM) learning session. First, in DRM, two subnetworks, i.e. domain-private network and domain-agnostic network are specially designed for learning modality features and identity features, respectively.
Second, in FFM, the identity features are fused with domain features to achieve cross-modal bi-directional identity feature transformation, which, to a large extent, further disentangles the modality information and identity information. Third, considering that the distribution imbalance between easy and hard pairs exists in cross-modal datasets, which increases the risk of model bias, the identity preserving guided metric learning with adaptive hard pairs penalization is proposed in our FAN. The proposed APM also guarantees the cross-modality intra-class compactness and inter-class separation. Extensive experiments on benchmark cross-modal face datasets show that our FAN outperforms SOTA methods.

\end{abstract}

\section{Introduction}
Heterogeneous face recognition (HFR) aims to solve the matching problem of face images with different modalities, which has attracted much attention and impressive progress in biometrics and computer vision community.

Early works in HFR focus on the domain-agnostic feature extraction and metric learning. With limited training data, hand-crafted features are focused by learning effective distance metrics. However, these methods still face unsatisfactory performance and a bottleneck due to the limited representation power of the hand-crafted features.

Recently, deep convolutional neural networks (CNN) have made impressive progress in HFR tasks. The potential of deep networks for tackling the HFR bottleneck is shown. Although the performance of the deep CNN models is much better than traditional ones, HFR still remains to be challenging due to some reasons. First, the NIR and VIS images are captured using different sensors, and they have large modality differences which seriously attack the feature representation. Therefore, learning domain-agnostic features in heterogeneous face images is a challenging task. Second, most of existing HFR datasets are of small-scale, which prohibits the effective training of deep networks. So, over-fitting during network training easily happens, and it is incapable of representing identity discriminative features.

In this paper, we aim to solve the two aforementioned problems: domain private and agnostic feature representation and network over-fitting towards model bias in HFR applications. Specifically, we propose a new Feature Aggregation Network (FAN), which mainly contains three sessions: Disentangled Representation Module (DRM), Feature Fusion Module (FFM) and Adaptive Penalty Metric(APM) Learning. Different from the general face recognition, in HFR the facial appearance contains both domain private and agnostic information, which, however, received little attention to disentangle and characterize the two kinds of information. In order to solve this problem, we propose a Disentangled Representation Module (DRM), which is designed with two subnetworks, \emph{i.e.} domain-agnostic network and domain-private network in Siamese structure. The domain-agnostic network and domain-private network can be trained by identity classification and domain classification, respectively, for disentangling the identity features and domain features partially.

Nevertheless, there are many redundant information still remaining in the identity features extracted from DRM, and it may shrink the discriminative ability of identity features. In order to further disentangle the feature representation, we propose a Feature Fusion Module (FFM), which fuses identity features and domain features through different ways to obtain four kinds of fusion features. The fusion features are then used to train the identity enhanced network for identity classification, such that the identity and modality information can be further disentangled.

Furthermore, considering the fact that in the cross-modal datasets, distribution imbalance between easy and hard pairs always exists, which increases the training over-fitting and model bias risk towards the larger side, the adaptive penalty metric(APM) learning with adaptive hard pairs self-penalization is proposed to train the FAN framework. Besides that, the proposed APM can also guarantee the identity features with smaller intra-class cross-modality distance and larger inter-class cross-modality distance.

\begin{figure*}
  \centering
  \includegraphics[width=1.00\textwidth]{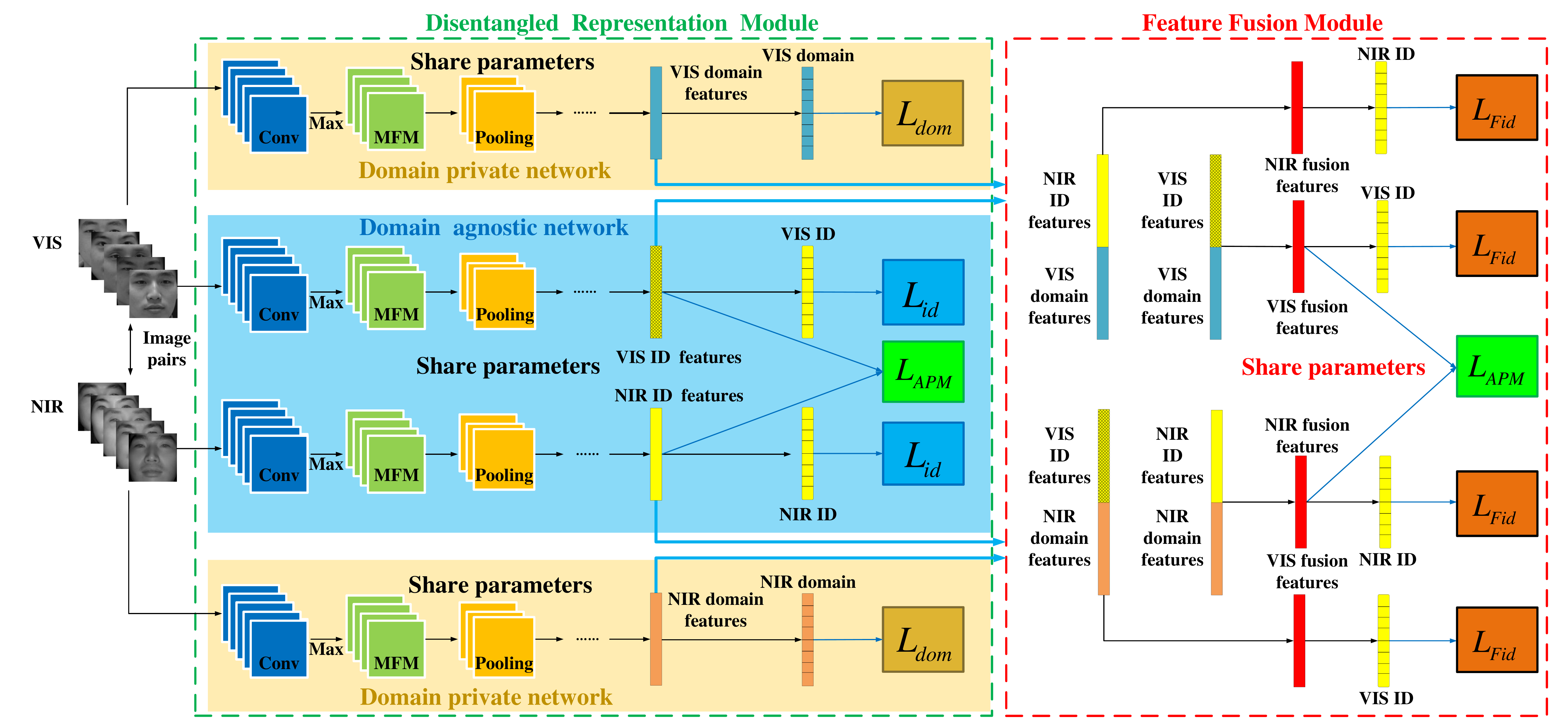}\\
  \caption{The architecture of the proposed FAN for VIS-NIR matching. Three important parts: Disentangled Representation Module (DRM), Feature Fusion Module (FFM) and the Adaptive Penalty Metric (APM) learning session are included. Where $L_{id}$, $L_{dom}$ and $L_{Fid}$ are cross-entrory loss.}\label{Fig1}
\end{figure*}
Due to the small-scale HFR dataset, too many parameters may easily lead to over-fitting, which encourages us to consider the parameter shared network structure. Hence, two pairs of Siamese networks are deployed as shown in Fig.~\ref{Fig1}.

In summary, our main contributions are as follows:

\begin{itemize}
\item A novel feature aggregation network (FAN) is proposed for learning modality-agnostic features, in which a novel disentangled representation module (DRM) containing two Siamese subnetworks (domain-private network and domain-agnostic network) is proposed. With the DRM, the modality and identity features can be partially disentangled.
\item  A feature fusion module (FFM) is proposed to further disentangle the identity information and the modality information by bi-directional identity feature transformation (cross-mapping) of the domain feature and identity feature.
\item  We propose an adaptive penalty metric (APM) learning session with adaptive hard pairs self-penalization, which, to a large extent, reduces the model bias during training phase. Simultaneously, it also guarantees cross-modality intra-class compactness and inter-class separation.
\item Our method achieves the state-of-the-art performance on three benchmark VIS-NIR databases, i.e. 99.6\% on CASIA NIR-VIS 2.0, 100.0\% on Oulu-CASIA NIR-VIS and 99.2\% on BUAA-VisNir in Rank-1 accuracy.
\end{itemize}
\section{Related Work}
\subsection{Heterogeneous Face Recognition (HFR)}
\noindent For HFR, the most important is how to eliminate the variations caused by modality difference. The methods can be categorized into three groups: subspace learning, modality-invariant feature learning and image synthesis.

\textbf{Subspace learning} maps the data from different modalities into a new common subspace, so that their modality difference become comparable and can be reduced. Lin \emph{et al} \cite{lin2006inter} proposed a common discriminant feature extraction (CDFE) method, which enhances intra-class consistency and sparseness inter-class of multi-modal data by learning a common feature space. Lei and Li \cite{lei2009coupled} proposed a spectral regression-based method to learn a discriminative subspace. Recently, the W-CNN proposed by He \emph{et al} \cite{he2018wasserstein} combines the subspace learning with CNN to obtain identity information and spectral information of modality difference.

\textbf{Modality-invariant feature learning} aims to extract modality-invariant features by removing the modality variations in extracting or learning face features, which is robust to modality changes. Most of these methods are hand-crafted features based, such as local binary patterns (LBP), difference-of-gaussian (DoG), histogram of oriented gradient (HoG). Huo \emph{et al.} \cite{huo2017heterogeneous} proposed a discriminative feature learning method, which maximizes the variance of the inter-class and minimizes the variance of the inter-class by learning some image filters.

\textbf{Image synthesis} maps the data from one modality into another by synthesis, and then learns the features to match. Liu \emph{et al} \cite{liu2005nonlinear} proposed a method based on local linear preserving to generate sketch image. Wang \emph{et al} use the multi-scale markov random field for photo-to-sketch synthesis and recognition. Lezama \emph{et al} \cite{lezama2017not} proposed a cross-spectral hallucination and low-rank embedding method to synthesize VIS images from NIR images for matching.  Recently, the Generative Adversarial Network (GAN) \cite{goodfellow2014generative} has drawn much attention in image synthesis field. Song \emph{et al} \cite{song2018adversarial} utilized a Cycle-GAN \cite{zhu2017unpaired} to realize a cross-spectral face hallucination, which improving the performance of HFR via generation.
\subsection{Distance Metric Learning}
\noindent The goal of metric learning is to learn a distance function to maximize the inter-class variations and minimize the intra-class variations. Previously, most works focus on learning Mahalanobis distance-based metrics, but which are based on a single modality and usually divided into local based and global based. Most of the previous methods are local-based. Weinberger and Saul \cite{weinberger2009distance} proposed a large margin nearest neighbor (LMNN) method to force a margin between a sample's nearest neighbors of the same class and its nearest neighbors of different classes. Since the Mahalanobis distance metric is developed for single modality, it is unable to remove variations across modalities. There have been many methods attempted to learn cross-modality metrics. In \cite{mignon2012cmml}, a cross-modal metric learning (CMML) method was proposed to learn metrics by using pairwise constraints. Kang \emph{et al} \cite{kang2015cross} proposed a low rank bilinear cross-modality similarity learning method, which used the logistic loss and pairwise distance constraints to learn a bilinear similarity function.

\subsection{Disentangled Representations}
\noindent Learning the feature representations from the data for each task wastes a lot of time and effort. The probability generation model provides a general framework for learning representations. This model consists of a joint probability distribution of data and potential random variables, which can be found by the posterior probability of the potential variable of the given data. Kingma \cite{kingma2013auto} proposed a Variational Auto-Encoder(VAE) framework to learn the reasoning model and generative model together, and infer latent variables on data. Recently, Luan Tran \emph{et al} \cite{tran2017disentangled} proposed Disentangled representation (DR)-GAN, in which the extracted image representation, noise and attribute codes are sent to the generator to generate a new pose-invariant face image for disentangling the pose variation and identity representation. Wu \emph{et al} \cite{wu2019disentangled} proposes the Disentangled Variational Representation (DVR) for cross-modal matching, which uses the variational lower bound to estimate the posterior and optimize the disentangled latent variable space.

\section{The Proposed FAN Approach}
\noindent In this section, we elaborate the design of our FAN model, including the Disentangled Representation Module (DRM), Feature Fusion Module (FFM) and Adaptive Penalty Metric (APM) loss.

\subsection{Disentangled Representation Module}
\noindent Firstly, due to the fact that the facial appearance contains both domain-agnostic (identity) information and domain-private (modality) information, which are highly entangled, the proposed DRM contains two Siamese subnetworks for disentangling the identity features and domain features from the original image. With the supervision of identity classification, the identity information is learned by the domain-agnostic network. The goal of domain-private network is learning the modality information that can preserve the characteristics of each domain by utilizing a domain classifier.

Specifically, the DRM takes a VIS-NIR image pair as input. Then in order to reduce the number of parameters and modality discrepancy, the parameter-shared structure is designed for the two Siamese subnetworks. As shown in Fig.~\ref{Fig1}, the DRM has four streams, with each stream the LightCNN~\cite{wu2018light} is used as backbone due to its superior recognition performance and light-weight parameters. Suppose the domain-agnostic network to be $f(*;\mathbf{\Theta})$ and the domain-private network $g(*;\mathbf{\Phi})$, where $\mathbf{\Theta}$ and $\mathbf{\Phi}$ are network parameters. The domain-agnostic network are initialized by the pre-trained models on large VIS dataset.

\indent We denote the VIS and NIR images as $I^{v}$ and $I^{n}$, respectively. For each VIS-NIR image pair ($I^{v}$, $I^{n}$), we can extract the identity features $\textbf{x}^{q}=f(I^{q};\mathbf{\Theta})$ and modality (domain) features $\textbf{m}^{q}=f(I^{q};\mathbf{\Phi})$, where $q\in \{ v, n \}$.

\indent The identity feature $\textbf{x}^{v}$ and $\textbf{x}^{n}$ are fed into a cross-entropy loss based identity classification layer for supervising the training step, which can be formulated as:
\begin{equation}\label{5}
  L_{id}= {\sum\limits_{q\in \{ v, n \}}\textit{\rm{CE}}}(\textbf{x}^{q},y^{q};\mathbf{\Theta}),
\end{equation}
where $y^{q}, q\in \{ v, n \}$ is the identity label and $\rm{CE}(\cdot)$ is the cross-entropy loss.

Then the domain features $\textbf{m}^{v}$ and $\textbf{m}^{n}$ are fed into a cross-entropy loss based domain classification layer, which is written as:
\begin{equation}\label{6}
  L_{dom}={\sum\limits_{q\in \{ v, n \}}\textit{\rm{CE}}}(\textbf{m}^{q},d^{q};\mathbf{\Phi}),
\end{equation}
where $d^{q}, q\in \{ v, n \}$ denotes the binary domain label (0 and 1 for two domains).

\indent The domain-agnostic network and domain-private network can be effectively trained by minimizing Eq.~\eqref{5} and Eq.~\eqref{6}, respectively, for disentangling the identity information and the modality information effectively.

\subsection{Feature Fusion Module}
\noindent As is shown in Fig.~\ref{Fig1}, the FFM includes a feature fusion part and an identity enhanced network. In order to further disentangle the identity and modality information, the identity features are concatenated with the domain features to form four domain-specified identity features as follow:
\begin{equation}\label{10}
\begin{aligned}
\textbf{s}^{q}=[\textbf{x}^{v}, \textbf{m}^{q}],\textbf{t}^{q}=[\textbf{x}^{n}, \textbf{m}^{q}] ,q\in \{ v, n \}
\end{aligned}
\end{equation}
We think that domain-specified identity features contain enough information, which can reconstruct a domain-specified face image. So the process that the identity features are fused with different domain features are treated as the cross-modal bi-directional identity feature transformation.

Then four kinds of new domain-specified identity features are fed into the identity enhanced network for identity classification. Since these new domain-specified identity features contains both modality information and identity information, by focusing on the discrimination of identity information in the identity enhanced network, the ability of disentangling the identity features of the domain-agnostic network can be improved.

Specifically, for each VIS-NIR image pair ($I^{v}$, $I^{n}$) and the corresponding identity label($y^{v}$, $y^{n}$), we utilize the DRM to extract the identity features and domain features ($\textbf{x}^{v}$, $\textbf{m}^{v}$, $\textbf{x}^{n}$, $\textbf{m}^{n}$) of VIS and NIR image, respectively. Then FFM take ($\textbf{x}^{v}$, $\textbf{m}^{v}$, $\textbf{x}^{n}$, $\textbf{m}^{n}$) as the input to formulate four kinds of fusion features $\textbf{s}^{q}=[\textbf{x}^{v}, \textbf{m}^{q}]$ and $\textbf{t}^{q}=[\textbf{x}^{n}, \textbf{m}^{q}]$, where $q\in \{ v, n \}$. The fusion features ($\textbf{s}^{v}$, $\textbf{s}^{n}$) and ($\textbf{t}^{v}$, $\textbf{t}^{n}$) share the same identity but different modalities.

\indent For identity discrimination, the four fusion features ($\textbf{s}^{v}$, $\textbf{s}^{n}$, $\textbf{t}^{v}$, $\textbf{t}^{n}$) are fed into the identity enhanced network, which includes two fully-connected layers with parameters $\textbf{W}$. The dimension of the fusion feature is reduced by the first fully-connected layer, and the second fully-connected layer is used for identity classification. We employ the cross-entropy loss based identity classification layer for supervising the training step as follows:

\begin{equation}\label{11}
\begin{aligned}
  L_{Fid}= &{\sum\limits_{q\in \{ v, n \}}\textbf{\rm{CE}}}(\textbf{s}^{q}, y^{v}; \textbf{W})
  +{\sum\limits_{q\in \{ v, n \}}\textbf{\rm{CE}}}(\textbf{t}^{q}, y^{n}; \textbf{W}),
\end{aligned}
\end{equation}

With the supervision of the identity classification, the more discriminative identity information can be extracted by the domain-agnostic network, which further disentangles the modality information implied in the identity features.

\begin{figure*}[htpb]
  \centering
  \includegraphics[width=1.0\textwidth]{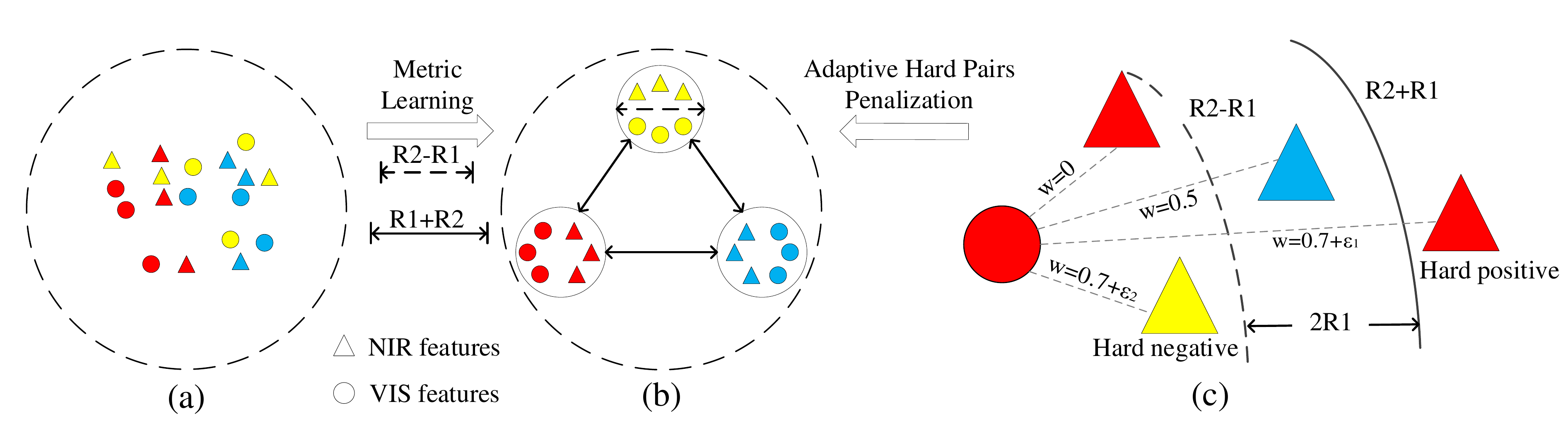}\\
  \caption{Illustration of the proposed APM learning idea. Two merits are included: (1) large-margin metric learning for smaller intra-class cross-modal pairwise distances and larger inter-class cross-modal pairwise distances; (2) adaptive hard-pairs penalties guided large-margin metric learning for reducing model bias towards hard pairs. (a) Original identity feature space. (b) Ideal identity feature space. (c) Illustration of hard sample pairs, where $w$ denotes the loss weight, ${\varepsilon}_{1}$ is the adaptive penalty of the hard positives and ${\varepsilon}_{2}$ is the adaptive penalty of the hard negatives.}\label{Fig2}
\end{figure*}

\subsection{Adaptive Penalty Metric Learning}
\noindent \textbf{Large Margin based Metric (LMM)}. In HFR, the extracted facial features of different modalities usually lie in two separated spaces due to the large modality discrepancy. In this case, the distances of intra-class cross-modal pairs and inter-class cross-modal pairs are inseparable, as shown in Fig.~\ref{Fig2}\textcolor{red}{(a)}. Therefore, it is necessary to design a suitable and efficient metric that is able to adapt the modality variation, such that the distances of intra-class cross-modal pairs and inter-class cross-modal pairs can be separated in the metric space.

Fig.~\ref{Fig2}\textcolor{red}{(b)} shows the desired results by using large margin metric (LMM) loss, which focuses on the learning of the identity feature representation with two distance constraints, i.e. small intra-class cross-modal distances and large inter-class cross-modal distances. For each VIS-NIR image pair ($I^{v}$, $I^{n}$) with the corresponding identity ($y^{v}$, $y^{n}$), we suppose the extracted identity features to be ($\textbf{k}_{v}$, $\textbf{k}_{n}$). The pair-wise cross-modality feature distance is characterized by Euclidean distance, and can be formulated as:

\begin{equation}\label{12}
  D(\textbf{k}_{v}, \textbf{k}_{n})=||\textbf{k}_{v}-\textbf{k}_{n}||_{2},
\end{equation}

We define two distance thresholds $R_1$ and $R_2$ ($R_2>R_1$). It is worth noting that we use Euclidean distance to represent the distribution of the datasets, so it is reasonable that the setting of $R_1$ and $R_2$ should be task-specific. Additionally, due to the large distribution difference of the HFR data, we normalize the feature vector to partially eliminate the distribution difference. If $\textbf{k}_{v}$ and $\textbf{k}_{n}$ are from the same identity, the cross-modal distance $D(\textbf{k}_{v}, \textbf{k}_{n})$ is smaller than lower-margin $R_2-R_1$. Otherwise, the cross-modal distances $D(\textbf{k}_{v}, \textbf{k}_{n})$ of the different identity is bigger than upper-margin $R_2+R_1$. The formulation is represented as:

\begin{equation}\label{13}
{}\left\{ {\begin{array}{*{20}{l}}
{D(\textbf{k}_{v}, \textbf{k}_{n})^{2}< R_2-R_1,}&{y^{v}=y^{n}},\\
{D(\textbf{k}_{v}, \textbf{k}_{n})^{2}> R_2+R_1,}&{y^{v}\neq y^{n}},
\end{array}} \right.
\end{equation}

Specifically, we can simplify the constraint in Eq.\eqref{13} as:
\begin{equation}\label{14}
  \min_{\Theta} R_1-y_{vn}\big(R_2-D(\textbf{k}_{v}, \textbf{k}_{n})^{2}\big),
\end{equation}
where $\Theta$ is the model and $y_{vn}$ is the label indicator for the pair-wise data ($I_{V},I_{N}$), which is defined as:

\begin{equation}\label{15}
y_{vn}{\rm{=}}\left\{ {\begin{array}{*{20}{l}}
{\rm{+1,}}&{y^{v}=y^{n}}\\
{\rm{-1,}}&{y^{v}\neq y^{n}},
\end{array}} \right.
\end{equation}

\indent In order to speed up the training, the hinge loss function $h(x)=\max(0,x)$ is introduced into Eq.\eqref{14}, and there is:
\begin{equation}\label{16}
  \min_{\Theta} h(R_1-y_{vn}(R_2-D(\textbf{k}_{v}, \textbf{k}_{n})^{2})),
\end{equation}
\indent By using the hinge loss, the model only optimizes the sample pairs that violate the constraint for improving the training efficiency.

\textbf{Adaptive Penalty Metric (APM)}. However, the LMM neglects the fact that there is a serious sample imbalance in HFR data, and a forthcoming problem is the hard sample pairs may deteriorate the model training towards model bias. As shown in Fig.~\ref{Fig2}\textcolor{red}{(c)}, the sample pairs of different identities with distances smaller than the lower-margin are denoted as the hard negative pairs and the sample pairs of the same identity with distances bigger than the up-margin are set as the hard positive pairs. Therefore, the hard threshold $R_1$ and $R_2$ may not be adaptive.
On the basis of LMM, we further originally propose an Adaptive Penalty Metric (APM), which introduces self-adaptive penalty factors to the hard sample pairs, so as to improve the training quality of FAN model.

Let $\beta$ be the indicator label of the hard sample pairs, there is:
\begin{equation}\label{17}
{\beta \rm{=}}\left\{ {\begin{array}{*{20}{l}}
{\rm{1,}}&{if\ D(\textbf{k}_{v}, \textbf{k}_{n})^{2}>R_2+R_1,y_{vn}=1},\\
{\rm{1,}}&{if\ D(\textbf{k}_{v}, \textbf{k}_{n})^{2}<R_2-R_1,y_{vn}=-1},\\
{\rm{0,}}&{else},
\end{array}} \right.
\end{equation}
Since the penalty values of the hard positive and hard negative pairs are different, we denote the indicator label $\lambda=0$ for positives and $\lambda=1$ for negatives. Let $\varepsilon_1=\delta\cdot D(k_{V}, k_{N})^{2}$ and $\varepsilon_2=(1-\delta)D(k_{V}, k_{N})^{2}$ be the penalty of hard positives and hard negatives, respectively, where $\delta$ is the penalty trade-off parameters and $0<\delta<1$. The penalties ${\varepsilon}_{1}$, ${\varepsilon}_{2}$ enable the distance of hard positive pairs smaller and the distance of hard negative pairs larger. Then, Eq.~(\ref{13}) can be written as:
\begin{equation}\label{18}
{}\left\{ {\begin{array}{*{20}{l}}
{D(\textbf{k}_{v}, \textbf{k}_{n})^{2}+ {\varepsilon}_{1} < R_2-R_1,}&{\lambda=0},\\
{D(\textbf{k}_{v}, \textbf{k}_{n})^{2}- {\varepsilon}_{2} > R_2+R_1,}&{\lambda=1},
\end{array}} \right.
\end{equation}
Based on Eq.~(\ref{16}), the proposed APM loss is presented as:
\begin{equation}\label{19}
\begin{aligned}
  L_{APM}(\textbf{k}_{v}, \textbf{k}_{n})&= \max(R_1+\beta\varepsilon_{1}^{1-\lambda}\varepsilon_{2}^{\lambda}\\
  &-y_{vn}(R_2-D(\textbf{k}_{v}, \textbf{k}_{n})^{2}),0),
\end{aligned}
\end{equation}
\indent During the optimization process, the sample pairs of the same identity that violate this constraint will be pulled into the lower-margin of the anchor sample, and the pairs of different identities that violate the constraint will be pushed out to the upper-margin of the anchor sample. Under the supervision of the APM loss, the discrepancy of the the modality variation will be reduced, which thereby enhances the discrimination of identity features.\\
\indent We integrate the APM loss into the domain-agnostic network and FFM for model training. By adding the APM loss into the Eq.\eqref{5} and Eq.\eqref{11}, then there is:
\begin{equation}\label{20}
  L_{DA}= L_{id}+\alpha_{1}L_{APM}(\textbf{x}^{v}, \textbf{x}^{n}),
\end{equation}
\begin{equation}\label{21}
  L_{FFM}= L_{Fid}+ \alpha_{2}L_{APM}(\textbf{s}^{v},\textbf{t}^{n}).
\end{equation}
\noindent where $\alpha_{1}$ and $\alpha_{2}$ are the trade-off parameters.
\subsection{Training and Testing}
\noindent There are three stages in training phase. First, the parameters of the domain-agnostic network $\mathbf{\Theta}$ is initialized by a pre-trained model. Then, the Stochastic gradient descent (SGD) is used to optimize the domain-agnostic network until convergence by Eq.\eqref{20}.

Second, we initialize the parameters of the domain-private network $\mathbf{\Phi}$ randomly and employ SGD to optimize the domain-private network until convergence by Eq.\eqref{6}.

Finally, the identity features and domain features are fused to generate new domain-specified identity features in the feature fusion module. Then, we feed the new identity feature into the fusion feature layer and train the whole network until convergence by Eq.\eqref{21}. The optimization details are summarized in Algorithm ~\ref{algorithm1}.

\indent In testing phase, only the domain-agnostic network $f(*;\mathbf{\Theta})$ is used to obtain feature representations. Note that the domain-private network $f(*;\mathbf{\Phi})$ and FFM are not utilized for testing. Then, the cosine similarity is computed to decide whether the two given face images are of the same person or not.
\vspace{-0.1cm}
\begin{algorithm}[h]
\caption{Feature Aggregation Networks (FAN) Training. }
\label{algorithm1}
\hspace*{-0.05in} {\bf Input}:Training set: VIS images $I^{v}$, NIR images $I^{n}$, distance thresholds $R_1$, $R_2$, trade-off parameters $\alpha_{1}$ and $\alpha_{2}$.\\
\hspace*{-0.05in} {\bf Output:}The parameters of the domain-agnostic network $\mathbf{\Theta}$, domain-private network $\mathbf{\Phi}$ and FFM network $\textbf{W}$. 
\begin{algorithmic}[1]
\State Initialize $\mathbf{\Theta}$ by pre-trained model;
\While{not converged}
　　\State Obtain $\textbf{x}_{v}=f(I^{v};\mathbf{\Theta})$, $\textbf{x}_{n}=f(I^{n};\mathbf{\Theta})$;
　　\State Compute $L_{DA}$ via Eq.\eqref{20};
　　\State Update $\mathbf{\Theta}$ by SGD;
\EndWhile
\State Initialize $\mathbf{\Phi}$, $\textbf{W}$ randomly;
\While{not converged}
　　\State Obtain $\textbf{m}_{v}=g(I^{v};\mathbf{\Phi})$, $\textbf{m}_{n}=g(I^{n};\mathbf{\Phi})$;
　　\State Compute $L_{dom}$ via Eq.\eqref{6};
　　\State Update $\mathbf{\Phi}$ by SGD;
\EndWhile

\For{ t = 1, \ldots,T}
　　\State Obtain $\textbf{x}^{v}=f(I^{v};\mathbf{\Theta})$, $\textbf{x}^{n}=f(I^{n};\mathbf{\Theta})$;
　　\State Obtain $\textbf{m}_{v}=g(I^{v};\mathbf{\Phi})$, $\textbf{m}_{n}=g(I^{n};\mathbf{\Phi})$;
　　\State obtain fusion feature $\textbf{s}^{v}$ ,$\textbf{s}^{n}$, $\textbf{t}^{v}$, $\textbf{t}^{n}$;
　　\State Compute $L_{FFM}$ via Eq.\eqref{21};
　　\State Update $\mathbf{\Theta}$, $\mathbf{\Phi}$, $\textbf{W}$ by SGD;
\EndFor{\textbf{end for}}
\State \Return $\mathbf{\Theta}$, $\mathbf{\Phi}$, $\textbf{W}$
\end{algorithmic}
\end{algorithm}

\section{Experiments}
\subsection{Datasets and Protocol}
\noindent \textbf{The CASIA NIR-VIS 2.0 face Database} \cite{li2013casia} is widely used for evaluating the VIS to NIR face image recognition. It contains 17,580 images of 725 subjects with variations in pose, age, and resolution, \emph{etc}. This database includes two views, \emph{i.e.} View 1 and View 2, which are used for parameter tuning and performance evaluation, respectively. In this paper, we follow the View 2 of the standard protocol defined in~\cite{li2013casia}. There are 10-fold experiments in View 2, and each fold contains training and testing lists, which are non-overlapped. There are about 6,100 NIR images and 2,500 VIS images from about 360 identities for training in each fold. In the testing phase, cross-view face verification is conducted between the gallery set and probe set in each testing fold. There are 358 VIS images from different 358 subjects in gallery set. That is, each subject has only one VIS image. The probe set contains over 6000 NIR images from the same 358 subjects. In our experiments, parameters are tuned on View 1 first, then the performance of our proposed approach is tested on View 2.

\begin{table*}[htpb]
\begin{center}
\small
\begin{tabular}{|l|cc|ccc|ccc|}
\hline
& \multicolumn{2}{c|}{CASIA NIR-VIS 2.0}& \multicolumn{3}{c|}{Oulu-CASIA NIR-VIS} & \multicolumn{3}{c|}{BUAA-VisNir}\\
\hline
Method &  Rank-1   &   FAR=0.1\% &  Rank-1       &  FAR=1\%   &   FAR=0.1\% &  Rank-1       &  FAR=1\%   &   FAR=0.1\%\\
\hline
KDSR\cite{huang2012regularized}& 37.5         & 9.3   & 66.9   & 56.1    & 31.9   & 83.0      & 86.8    &69.5\\
H2(LBP3)\cite{Shao2017Cross}   & 43.8         & 10.1   & 70.8   & 62.0    & 33.6   & 88.8     & 88.8    & 73.4\\
Recon.+UDP \cite{juefei2015nir} & 78.5$\pm$ 1.7         & 85.8 & -& -& -& -& -& -\\
Gabor+RBM \cite{yi2015shared}  & 86.2$\pm$ 1.0         & 81.3$\pm$ 1.8 & -& -& -& -& -& -\\
Gabor+JB \cite{chen2012bayesian}  & 89.5$\pm$ 0.8   & 83.2$\pm$ 1.0 & -& -& -& -& -& -\\
Gabor+HJB \cite{shi2017cross}  & 91.6$\pm$ 0.8   & 89.9$\pm$ 0.9 & -& -& -& -& -& -\\
IDNet \cite{reale2016seeing}  & 87.1$\pm$ 0.9   & 74.5  & -& -& -& -& -& -\\
Hallucination \cite{lezama2017not}& 89.6$\pm$ 0.9        & - & -& -& -& -& -& -\\
TRIVET  \cite{liu2016transferring}  & 95.7$\pm$ 0.5         & 91.0$\pm$ 1.3 & 92.2  & 67.9  & 33.6 & 93.9   &93.0  & 80.9\\
IDR \cite{he2017learning} & 97.3$\pm$ 0.4         & 95.7$\pm$ 0.7 & 94.3 & 73.4 & 46.2   & 94.3      & 93.4    & 84.7\\
ADFL \cite{song2018adversarial}  & 98.2$\pm$ 0.3        & 97.2$\pm$ 0.3 & 95.5  & 83.0 & 60.7   & 95.2      & 95.3    & 88.0\\
CDL \cite{wu2018coupled}  & 98.6$\pm$ 0.2         & 98.3$\pm$ 0.1 & 94.3  & 81.6 & 53.9   & 96.9      & 95.9    & 90.1\\
W-CNN \cite{he2018wasserstein}  & 98.7$\pm$ 0.3         & 98.4$\pm$ 0.4 & 98.0  & 81.5 & 54.6   & 97.4      & 96.0    & 91.9\\
RCN \cite{deng2019residual} & 99.3$\pm$ 0.2         & 98.7$\pm$ 0.2 & -& -& -& -& -& -\\
MC-CNN \cite{deng2019mutual}& 99.4$\pm$ 0.1         & 99.3$\pm$ 0.1 & -& -& -& -& -& -\\
\hline
Ours  & \textbf{99.6}$\pm$ 0.1   & \textbf{99.4}$\pm$ 0.1 & \textbf{100} & \textbf{95.5} & \textbf{87.5} & \textbf{99.2}  & \textbf{98.8}   & \textbf{97.8}\\
\hline
\par
\end{tabular}
\end{center}
\caption{Comparisons with other state-of-the-art HFR methods on the CASIA NIR-VIS 2.0 database, the Oulu-CASIA NIR-VIS database and the BUAA-VisNir database.}
\label{table:r1}
\end{table*}

\indent \textbf{The BUAA-VisNir Face Database} \cite{huang2012buaa} consists of data from 150 subjects with 40 images per subject, among which there are 13 VIS-NIR pairs and 14 VIS images in different illuminations. The 9 VIS-NIR face pairs per subject, correspond to 9 distinct poses or expressions: neutral-frontal, left-rotation, right-rotation, tilt-up, tilt-down, happiness, anger, sorrow and surprise in our experiments. 900 images (\emph{i.e.} 450 VIS-NIR face pairs) of the first 50 subjects are used as the train set and 1800 facial images (\emph{i.e.} 900 VIS-NIR face pairs) from the remaining 100 identities are used as the test set. In the testing phase, only one VIS image of each subject is selected in the gallery set (\emph{i.e.} 100 VIS face images in total) and all NIR images of each identity are used as the probe set (\emph{i.e.} 900 NIR images).

\indent \textbf{The Oulu-CASIA NIR-VIS Face Database} \cite{chen2009learning} consists of data from 80 subjects with 6 expression variations. Following the protocols in \cite{he2018wasserstein}, 8 VIS-NIR face pairs from each expression of each subject are randomly selected in this paper. That is, there are a total of 96 images per subject. 20 subjects are selected as the train set and the remaining 20 subjects are adopted as the test set. On the testing stage, all VIS images from the 20 subjects are used as the gallery set, and all the corresponding NIR images are constituted as the probe set. The Rank-1 face identification accuracy and verification rate (VR) @false accept rate(FAR)=0.1\% are employed to evaluate the performances of algorithms.
\subsection{Implementation Details}
\noindent The LightCNN-29 \cite{wu2018light}, which is pre-trained on the MS-Celeb-1M dataset \cite{guo2016ms} and CASIA-WebFace dataset \cite{yi2014learning}, is employed as the backbone networks of our FAN method. MTCNN \cite{zhang2016joint} is also adopted to detect the landmarks of facial images, then align and crop the original RGB face images to gray-scale \(128\times128\) facial images.

\indent There are three steps in training phase. Firstly, we train the domain-agnostic network in DRM. The batch size is set to 64, and the initial learning rate is set 1e-4. Then, after 30 epochs, the domain-private network of DRM is trained with the initial learning rate 1e-3, and the batch size is set as 40. This training step is conducted about 5 epochs. Based on the former two steps, the whole network is trained end-to-end with the same initial learning rate and batch size of the $2^{\emph{nd}}$ step. The SGD is employed to optimize the parameters of our FAN. The momentum and weight decay are set as 0.9 and 2e-4, respectively. The trade-off parameters $\alpha_{1}$ and $\alpha_{2}$ are set equal to 0.01. The distance thresholds $R_1$ and $R_2$  are set to 5 and 100, respectively. The penalty parameter $\delta$ is set to 0.5.

\subsection{Comparison to the SOTA Models}
\noindent The comparison of our method with other state-of-the-art methods on the three HFR datasets is presented in Table ~\ref{table:r1}, which includes traditional methods and deep learning methods. Firstly, we compare our method with the traditional approaches, including KDSR\cite{huang2012regularized}, H2(LBP3)\cite{Shao2017Cross}, Recon.+UDP \cite{juefei2015nir}, Gabor+RBM \cite{yi2015shared}, Gabor+JB \cite{chen2012bayesian} and Gabor+HJB \cite{shi2017cross}. It can be seen that the performance of our method outperforms those shallow methods with a large margin. It shows that the superiority of deep learning based method to the traditional learning methods in HFR.

Secondly, the experimental results of FFN are also compared with state-of-the-art deep learning based methods for NIR-VIS face recognition, such as IDNet \cite{reale2016seeing}, Hallucination \cite{lezama2017not}, TRIVET \cite{liu2016transferring}, IDR \cite{he2017learning}, ADFL \cite{song2018adversarial}, CDL \cite{wu2018coupled}, W-CNN \cite{he2018wasserstein}, RCN \cite{deng2019residual} and MC-CNN\cite{deng2019mutual}. It is obvious that the recognition accuracies of our method are still higher than other deep learning methods. The experimental results demonstrate that the identity discrimination of the extracted features can be improved by disentangling the identity information and modality-specific information effectively.

The performance of our method outperforms other state-of-the-art methods. The results of our proposed approach can achieve 99.6\% on Rank-1, 99.4\% on VR@FAR=0.1\% on CASIA NIR-VIS 2.0 dataset. For Oulu-CASIA NIR-VIS dataset, FAN obtains 100\% on Rank-1, 95.5\% on VR@FAR=1\% and 87.5\% on VR@FAR=0.1\%. In BUAA-VisNir dataset, our method achieve 99.2\% on Rank-1, 98.8\% on VR@FAR=1\% and 97.8\% on VR@FAR=0.1\%, which are better than the previous state-of-the-art methods. The experimental results show that the identity discrimination of the extracted features can be improved by disentangling the identity information and modality-specific information effectively.
\begin{table*}[htpb]
\begin{center}
\fontsize{8}{12}\selectfont
\begin{tabular}{|l|cc|ccc|ccc|}
\hline
\multicolumn{1}{|l|}{} & \multicolumn{2}{c|}{\fontsize{5}{8} CASIA NIR-VIS 2.0} & \multicolumn{3}{c|}{ Oulu-CASIA NIR-VIS} & \multicolumn{3}{c|}{ BUAA-VisNir} \\
\hline
Method &  Rank-1       &  FAR=0.1\%   &  Rank-1       &  FAR=1\% &   FAR=0.1\%&  Rank-1       &  FAR=1\%   &   FAR=0.1\%\\
\hline
(a): Fine-tuned LightCNN-29 &97.8    & 97.0     & 99.0   &  85.1   &72.1   & 94.2     & 95.7    &90.0   \\
(b): (a)+LMM loss &99.1              &98.6     &100    &94.6     &79.5    &98.2       &98.0     &96.7   \\
(c): (a)+APM loss &99.3              & 98.9    & 100   & 95.3    &  83.9  & 98.4      & 98.1    & 97.0   \\
(d): (a)+FFM &99.2              & 98.9     &100    &94.9     &83.1    &98.3       & 97.9   & 96.9   \\
(e): (a)+LMM loss+FFM & 99.3             &99.1      &100    &95.4     &84.9    &98.9      &98.4    &97.2    \\
(f): (a)+APM loss+FFM &\textbf{99.6} & \textbf{99.4}     & \textbf{100}  & \textbf{95.5} &\textbf{87.5}  & \textbf{99.2}  & \textbf{98.8}   & \textbf{97.8}   \\
\hline
\par
\end{tabular}
\end{center}
\caption{Ablation study of the proposed FAN approach.}
\label{table22}
\end{table*}
\begin{figure*}[htpb]
  \centering
  \includegraphics[width=0.95\textwidth]{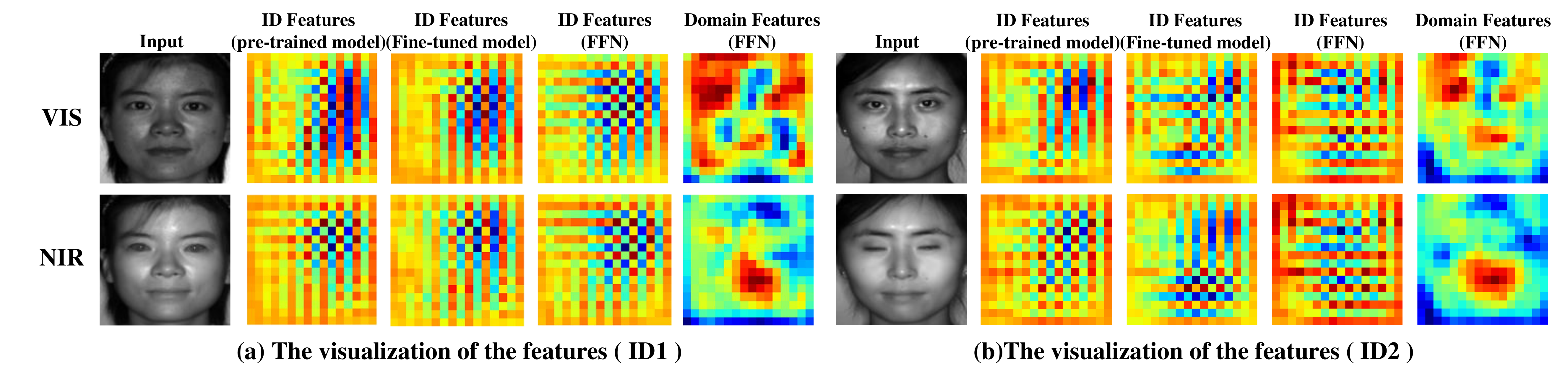}\\
  \caption{Visualization of identity and domain features. The red color represents high response while the blue color represents low response.}\label{Fig4}
 \vspace{-0.4cm}
\end{figure*}
\subsection{Ablation Study}
\noindent In order to illustrate the effectiveness of the key components of our proposed FAN approach, the ablation analysis is conducted. As presented in Table~\ref{table22}, we investigate four methods, including FAN (a)-(f). Compared with the FAN (a) and FAN (c), we observe that after fine-tuning the LightCNN-29 pre-trained model with the proposed APM loss, the performance on the VR@FAR=0.1\% can improved from 97.0\% to 98.9\% for CASIA NIR-VIS 2.0, from 72.1\% to 83.9\% for Oulu-CASIA NIR-VIS and from 90.0\% to 97.0\% for BUAA-VisNir, respectively. The results demonstrate that the proposed APM loss can effectively reduce the modality discrepancy, which thereby enhances the discrimination of identity features. Additionally, from the results between FAN (b) and FAN (c) and the results between FAN (e) and FAN (f), the superiority of APM over LMM by introducing adaptive hard-pair penalization is proved.

The comparison results of FAN (a) and FAN (d) show that the proposed FFM can improve the performance on the VR@FAR=0.1\%, and achieves 1.9\%, 9.0\% and 6.9\% improvements for CASIA NIR-VIS 2.0, Oulu-CASIA NIR-VIS and BUAA-VisNir, respectively. The results indicate that FFM can significantly disentangle the modality variations and thus reduce the modality discrepancy.

As shown in Table ~\ref{table22}, the FAN (f) shows the best performance, which achieves 99.4\%, 87.5\% and 97.8\% on VR@FAR=0.1\% for CASIA NIR-VIS 2.0, Oulu-CASIA NIR-VIS and BUAA-VisNir, respectively. The results verify that the proposed APM loss and FFM are useful, and our FAN method can effectively learn domain-agnostic and identity discriminative features.

\subsection{Feature Visualization Analysis}
\noindent To examine the effectiveness of our proposed FAN, we visualize the high-level domain features extracted by FAN and the high-level identity features from three models, including the pre-trained LightCNN-29 model, fine-tuned LightCNN-29 model and FAN. Fig.~\ref{Fig4} illustrates the visualization results of the ID features and domain features from the different identity. Obviously, due to the deeper domain layers contain more modality-specific information, the high-level domain features from different domains show large difference. Besides, the deeper ID layers have more identity information so that the high-level identity features are separated between different IDs. In addition, we observe that the ID features extracted by FAN model are much more similar than the feature extracted by pre-trained and fine-tuned model. Consequently, the above observations can show that our FAN can effectively extract domain information and identity information and deal with modality discrepancy.
\section{Conclusion}
\noindent In this paper, a Feature Aggregation Network (FAN) consisting of a disentanglement representation module (DRM), feature fusion module (FFM) and adaptive penalty metric (APM) learning is proposed for modality adaptive face recognition. The DRM is deployed to learn modality and identity features by using a domain-private subnetwork and a domain-agnostic subnetwork, respectively. In FFM, the identity feature are fused with domain feature to achieve bi-directional identity feature transformation for further domain-identity disentanglement. For training the FAN model, we propose a large margin based metric which enables the smaller intra-class cross-domain pairwise distances and larger inter-class cross-domain pairwise distances. Considering that in HFR data the easy and hard pairs are seriously imbalanced, we originally propose an adaptive hard-pair self-penalization on the large margin metric. The proposed APM effectively alleviates the model bias risk. Experiments on three benchmark NIR-VIS face recognition datasets demonstrate that our method outperforms the recent state-of-the-art models.

%
%
\section*{Acknowledgements}

\noindent This work was supported by the National Science Fund of China under Grants (61771079) and Chongqing Youth Talent Program.

{\small
\bibliographystyle{ieee}
\bibliography{references}
}
\end{document}